\documentclass[10pt,twocolumn,letterpaper]{article}

\usepackage{cvpr}
\usepackage{times}
\usepackage{epsfig}
\usepackage{graphicx}
\usepackage{amsmath}
\usepackage{amssymb}

\usepackage{algorithm}
\usepackage{algpseudocode}
\algrenewcommand\algorithmicindent{.9em}%

\usepackage{bbm}
\usepackage{comment}
\usepackage[british,UKenglish,USenglish,english,american]{babel}
\usepackage{multirow}
\newcommand{\tabincell}[2]{\begin{tabular}{@{}#1@{}}#2\end{tabular}}

\usepackage[pagebackref=true,breaklinks=true,letterpaper=true,colorlinks,bookmarks=false]{hyperref}

\cvprfinalcopy 

\ifcvprfinal\fi

\begin{document}
	
	\title{Towards High Performance Video Object Detection}
	
	\author{Xizhou Zhu\thanks{This work is done when Xizhou Zhu is intern at Microsoft Research Asia} \qquad Jifeng Dai \qquad Lu Yuan \qquad Yichen Wei \vspace{8pt}\\
		Microsoft Research Asia\\
		\hspace{0.7in}{\tt\small \{v-xizzhu,jifdai,luyuan,yichenw\}@microsoft.com} \\
	}
	
	\maketitle
	\thispagestyle{empty}
	
	\begin{abstract}
		There has been significant progresses for image object detection in recent years. Nevertheless, video object detection has received little attention, although it is more challenging and more important in practical scenarios.
		
		Built upon the recent works~\cite{zhu2016dff,zhu2017flow}, this work proposes a unified approach based on the principle of multi-frame end-to-end learning of features and cross-frame motion. Our approach extends prior works with three new techniques and steadily pushes forward the performance envelope  (speed-accuracy tradeoff), towards high performance video object detection.
	\end{abstract}
	
	\section{Introduction}
	
	Recent years have witnessed significant progress in object detection~\cite{huang2016speed} in still images.
	However, directly applying these detectors to videos faces new challenges. First, applying the deep networks on all video frames introduces unaffordable computational cost. Second, recognition accuracy suffers from deteriorated appearances in videos that are seldom observed in still images, such as motion blur, video defocus, rare poses, etc.
	
	There has been few works on video object detection. The recent works~\cite{zhu2016dff,zhu2017flow} suggest that principled multi-frame end-to-end learning is effective towards addressing above challenges. Specifically, data redundancy between consecutive frames is exploited in~\cite{zhu2016dff} to reduce the expensive feature computation on most frames and improve the speed. Temporal feature aggregation is performed in~\cite{zhu2017flow} to improve the feature quality and recognition accuracy. These works are the foundation of the ImageNet Video Object Detection Challenge 2017 winner~\cite{deng2017vid}.
	
	The two works focus on different aspects and presents their own drawbacks. \emph{Sparse feature propagation} (see Eq.~\eqref{eq.feature_propagation})  is used in~\cite{zhu2016dff} to save expensive feature computation on most frames. Features on these frames are propagated from sparse key frames cheaply. The propagated features, however, are only approximated and error-prone, thus hurting the recognition accuracy. Multi-frame \emph{dense feature aggregation} (see Eq.~\eqref{eq.feature_aggregation}) is performed in~\cite{zhu2017flow} to improve feature quality on all frames and detection accuracy as well. Nevertheless, it is much slower due to repeated motion estimation, feature propagation and aggregation.
	
	The two works are complementary in nature. They also share the same principles: motion estimation module is built into the network architecture and end-to-end learning of all modules is performed over multiple frames. 
	
	Built on these progresses and principles, this work presents a unified approach that is faster, more accurate, and more flexible. Specifically, three new techniques are proposed. First, \emph{sparsely recursive feature aggregation} is used to retain the feature quality from aggregation but as well reduce the computational cost by operating only on sparse key frames. This technique combines the merits of both works~\cite{zhu2016dff,zhu2017flow} and performs better than both.
	
	Second, \emph{spatially-adaptive partial feature updating} is introduced to recompute features on non-key frames wherever propagated features have bad quality. The feature quality is learnt via a novel formulation in the end-to-end training. This technique further improves the recognition accuracy.
	
	Last, \emph{temporally-adaptive key frame scheduling} replaces the previous fixed key frame scheduling. It predicts the usage of a key frame accordingly to the predicted feature quality above. It makes the key frame usage more efficient.
	
	The proposed techniques are unified with the prior works~\cite{zhu2016dff,zhu2017flow} under a unified viewpoint. Comprehensive experiments show that the three techniques steadily pushes forward the performance (speed-accuracy trade-off) envelope, towards high performance video object detection. For example, we achieve $77.8\%$ mAP score at speed of 15.22 frame per second. It establishes the new state-of-the-art.
	
	\section{From Image to Video Object Detection}
	
	Object detection in static images has achieved significant progress in recent years using deep CNN~\cite{huang2016speed}. State-of-the-art detectors share the similar methodology and network architecture, consisting of two \emph{conceptual} steps.
	
	First step extracts a set of convolutional feature maps $F$ over the whole input image $I$ via a fully convolutional backbone network~\cite{simonyan2015very,szegedy2015going,he2016deep,szegedy2016inception,xie2017resnext,huang2016densely,chollet2016xception,howard2017mobilenets,zhang2017shufflenet}. The backbone network is usually pre-trained on the ImageNet classification task and fine-tuned later. In this work, it is called \emph{feature network}, $\mathcal{N}_{\text{feat}}(I)=F$. It is usually deep and slow. Computing it on all video frames is unaffordable.
	
	Second step generates detection result $y$ upon the feature maps $F$, by performing region classification and bounding box regression over either sparse object proposals~\cite{girshick2014rich,he2014spatial,girshick2015fast,ren2015faster,dai2016rfcn,lin2016fpn,he2017mask,dai2017deformable} or dense sliding windows~\cite{liu2016ssd,redmon2016you,redmon2016yolo9000,lin2017focal}, via a multi-branched sub-network. It is called \emph{detection network} in this work, $\mathcal{N}_{\text{det}}(F)=y$. It is randomly initialized and jointly trained with $\mathcal{N}_{\text{feat}}$. It is usually shallow and fast.
	
	\subsection{Revisiting Two Baseline Methods on Video}
	\label{sec.method_two_baselines}
	
	\paragraph{Sparse Feature Propagation~\cite{zhu2016dff}.} It introduces the concept of \emph{key frame} for video object detection, for the first time. The motivation is that similar appearance among adjacent frames usually results in similar features. It is therefore unnecessary to compute features on all frames.
	
	During inference, the expensive feature network $\mathcal{N}_{\text{feat}}$ is applied only on sparse key frames (\eg, every $10^{th}$). The feature maps on any non-key frame $i$ are propagated from its preceding key frame $k$ by per-pixel feature value warping and bilinear interpolation. The between frame pixel-wise motion is recorded in a two dimensional \emph{motion field} $M_{i \rightarrow k}$\footnote{Since the warping $\mathcal{W}$ from frame $k$ to $i$ adopts backward warping, we directly estimate and use backward motion field $M_{i \rightarrow k}$ for convenience.}. The propagation from key frame $k$ to frame $i$ is denoted as
	\begin{equation}
	F_{k \rightarrow i} = \mathcal{W}(F_{k}, M_{i \rightarrow k}),
	\label{eq.feature_propagation}
	\end{equation}
	where $\mathcal{W}$ represents the feature warping function. Then the detection network $\mathcal{N}_{\text{det}}$ works on $F_{k \rightarrow i}$, the approximation to the real feature $F_i$, instead of computing $F_i$ from $\mathcal{N}_{\text{feat}}$.
	
	The motion field is estimated by a lightweight flow network, $\mathcal{N}_{\text{flow}}(I_k, I_i) = M_{i \rightarrow k}$~\cite{dosovitskiy2015flownet}, which takes two frames $I_k, I_i$ as input. End-to-end training of all modules, including $\mathcal{N}_{\text{flow}}$, greatly boosts the detection accuracy and makes up for the inaccuracy caused by feature approximation. Compared to the single frame detector, because the computation of $\mathcal{N}_{\text{flow}}$ and Eq.~(\ref{eq.feature_propagation}) is much cheaper (dozens, see Table 2 in~\cite{zhu2016dff}) than feature extraction in $\mathcal{N}_{\text{feat}}$, method in~\cite{zhu2016dff} is much faster (up to $10\times$) with small accuracy drop (up to a few mAP points) (see, Figure 3 in~\cite{zhu2016dff}).
	
	\paragraph{Dense Feature Aggregation~\cite{zhu2017flow}.} It introduces the concept of \emph{temporal feature aggregation} for video object detection, for the first time. The motivation is that the deep features would be impaired by deteriorated appearance (\eg, motion blur, occlusion) on certain frames, but could be improved by aggregation from nearby frames.
	
	During inference, feature network $\mathcal{N}_{\text{feat}}$ is \emph{densely} evaluated on all frames. For any frame $i$, the feature maps of all the frames within a temporal window $[i-r, i+r]$ ($r = 2 \sim 12$ frames) are firstly warped onto the frame $i$ in the same way to~\cite{zhu2016dff} (see Eq.~(\ref{eq.feature_propagation})), forming a set of feature maps $\{F_{k\rightarrow i}| k \in [i-r, i+r]\}$. Different from \emph{sparse feature propagation}~\cite{zhu2016dff}, the propagation occurs at every frame instead of key frame only. In other words, every frame is viewed as key frame.
	
	The aggregated feature maps $\bar{F}_i$ at frame $i$ is then obtained as the weighted average of all such feature maps,
	\begin{equation}
	\bar{F}_i(p)=\sum_{k \in [i-r, i+r]} W_{k\rightarrow i}(p)\cdot F_{k\rightarrow i}(p), \forall p,
	\label{eq.feature_aggregation}
	\end{equation}
	where the weight $W_{k\rightarrow i}$ is adaptively computed as the similarity between the propagated feature maps $F_{k\rightarrow i}$ and the real feature maps $F_i$. Instead, the feature $F$ is projected into an embedding feature $F^e$ for similarity measure, and the projection can be implemented by a tiny fully convolutional network (see Section 3.4 in~\cite{zhu2017flow}).
	\begin{equation}
	W_{k\rightarrow i}(p) = \exp \left({\frac{F^e_{k\rightarrow i}(p)\cdot F^e_i(p)}{|F^e_{k\rightarrow i}(p)|\cdot|F^e_i(p)|}}\right), \forall p.
	\label{eq.aggregation_weight}
	\end{equation}
	Note that both Eq.~(\ref{eq.feature_aggregation}) and~(\ref{eq.aggregation_weight}) are in a position-wise manner, as indicated by enumerating the location $p$. The weight is normalized at every location $p$ over nearby frames, $\sum_{k \in [i-r, i+r]}W_{k\rightarrow i}(p)=1$.
	
	Similarly as~\cite{zhu2016dff}, all modules including the flow network and aggregation weight, etc., are jointly trained. Compared to the single frame detector, the aggregation in Eq.~(\ref{eq.feature_aggregation}) greatly enhances the features and improves the detection accuracy (about $3$ mAP points),  especially for the fast moving objects (about $6$ mAP points) (see Table 1 in~\cite{zhu2017flow}). However, runtime is about $3$ times slower due to the repeated flow estimation and feature aggregation over dense consecutive frames.
	
	\section{High Performance Video Object Detection}
	\label{sec.high_performance_detection}
	\begin{figure}
		\begin{center}
			\includegraphics[width=1.0\linewidth]{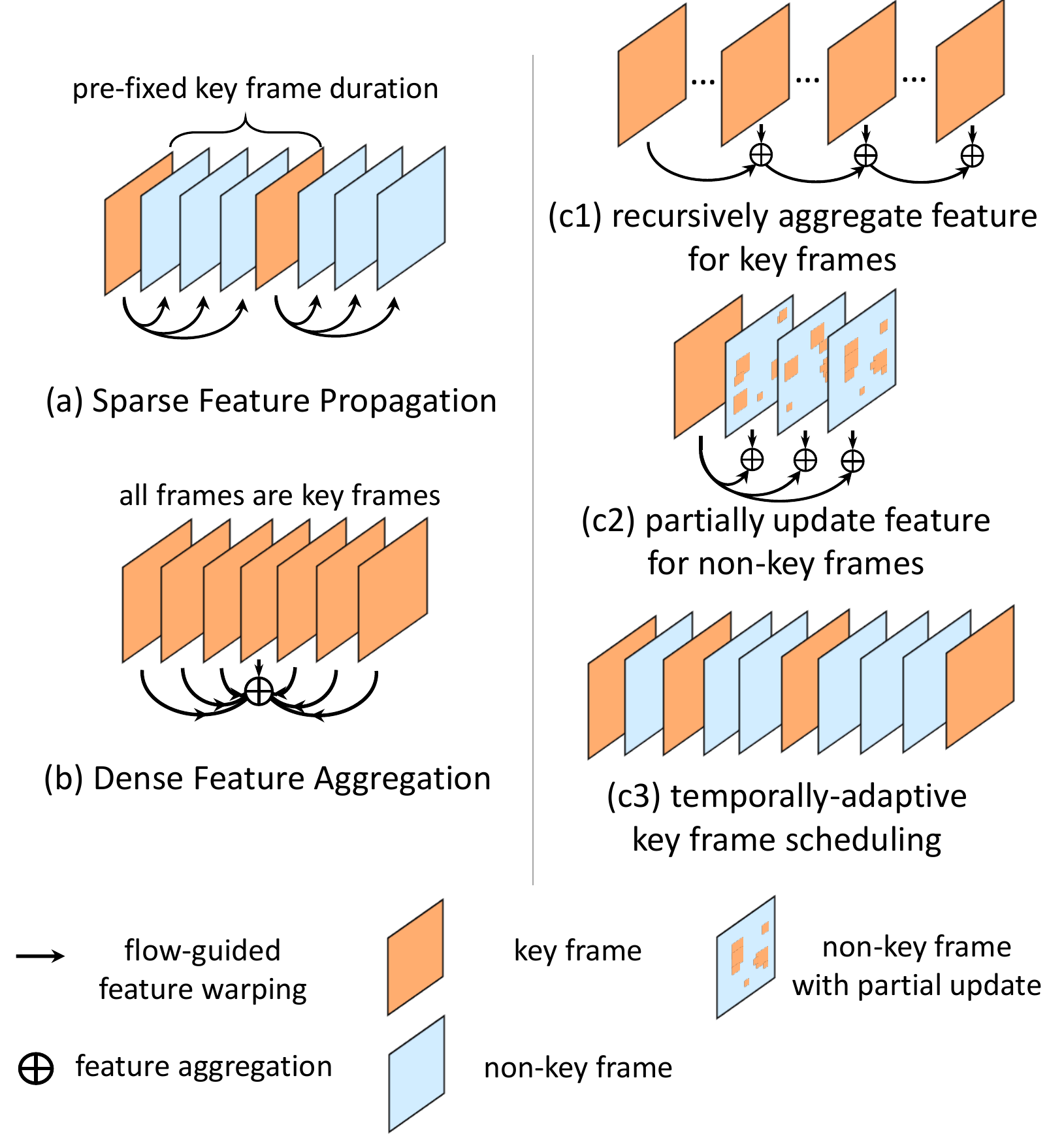}
		\end{center}
		\vspace{-0.5em}
		\caption{Illustration of the two baseline methods in~\cite{zhu2016dff,zhu2017flow} and three new techniques presented in Section~\ref{sec.high_performance_detection}.}
		\vspace{-0.5em}
		\label{fig.method_illustration}
	\end{figure}
	The difference between the above two methods is apparent. ~\cite{zhu2016dff} reduces feature computation by feature approximation, which decreases accuracy. ~\cite{zhu2017flow} improves feature quality by adaptive aggregation, which increases computation. They are naturally complementary.
	
	On the other hand, they are based on the same two principles: 1) \emph{motion estimation} module is indispensable for effective \emph{feature level} communication between frames; 2) \emph{end-to-end learning over multiple frames} of all modules is crucial for detection accuracy, as repeatedly verified in~\cite{zhu2016dff,zhu2017flow}.
	
	Based on the same underlying principles, this paper presents a common framework for high performance video object detection, as summarized in Section~\ref{sec.unified_viewpoint}. It proposes three novel techniques. The first (Section~\ref{sec.sparse_and_recursive}) exploits the complementary property and integrates the methods in~\cite{zhu2016dff,zhu2017flow}. It is both accurate and fast. The second (Section~\ref{sec.spatially_adaptive}) extends the idea of adaptive feature computation from temporal domain to spatial domain, resulting in spatially-adaptive feature computation that is more effective. The third (Section~\ref{sec.temporally_adaptive}) proposes adaptive key frame scheduling that further improves the efficiency of feature computation.
	
	These techniques are simple and intuitive. They naturally extend the previous works. Each one is built upon the previous one(s) and steadily pushes forward the performance (runtime-accuracy trade off) envelope, as verified by extensive experiments in Section~\ref{sec.experiment}.
	
	The two baseline methods and the three new techniques are illustrated in Figure~\ref{fig.method_illustration}.
	
	\subsection{Sparsely Recursive Feature Aggregation}
	\label{sec.sparse_and_recursive}
	
	Although Dense Feature Aggregation~\cite{zhu2017flow} achieves significant improvement on detection accuracy, it is quite slow. On one hand, it densely evaluates feature network $\mathcal{N}_{\text{feat}}$ on all frames, however that is unnecessary due to the similar appearance among adjacent frames. One the other hand, feature aggregation is performed on multiple feature maps and thus multiple flow fields are needed to be estimated, which largely slow down the detector.
	
	Here we propose \emph{Sparsely Recursive Feature Aggregation}, which both evaluates feature network $\mathcal{N}_{\text{feat}}$ and applies recursive feature aggregation only on sparse key frames. 
	Given two succeeding key frames $k$ and $k'$, the aggregated feature at frame $k'$ is computed by 
	\begin{equation}
	\bar{F}_{k'} = W_{k\rightarrow k'} \odot \bar{F}_{k \rightarrow k'} + W_{k'\rightarrow k'} \odot F_{k'},
	\label{eq.sparse_recursive_aggregation}
	\end{equation}
	where $\bar{F}_{k \rightarrow k'} = \mathcal{W}(\bar{F}_{k}, {M}_{k' \rightarrow k})$, and $\odot$ denotes element-wise multiplication. The weight is correspondingly normalized by $W_{k\rightarrow k'}(p)+W_{k'\rightarrow k'}(p)=1$ at every location $p$.
	
	This is a recursive version of Eq.~\eqref{eq.feature_aggregation}, and the aggregation only happens at sparse key frames. In principle, the aggregated key frame feature $\bar{F}_{k}$ aggregates the rich information from all history key frames, and is then propagated to the next key frame $k'$ for aggregating the original feature $F_{k'}$.
	
	\subsection{Spatially-adaptive Partial Feature Updating}
	\label{sec.spatially_adaptive}
	
	Although Sparse Feature Propagation~\cite{zhu2016dff} achieves remarkable speedup by approximating the real feature $F_{i}$, the propagated feature map $F_{k \rightarrow i}$ is error-prone due to some parts with changing appearance among adjacent frames.
	
	For non-key frames, we want to use the idea of feature propagation for efficient computation, however Eq.~\eqref{eq.feature_propagation} is subject to the quality of propagation.
	To quantify whether the propagated feature $F_{k \rightarrow i}$ is a good approximation of $F_{i}$, a feature temporal consistency ${Q}_{k\rightarrow i}$ is introduced. We add a sibling branch on the flow network $\mathcal{N}_{\text{flow}}$ for predicting $Q_{k \rightarrow i}$, together with motion field $M_{i \rightarrow k}$, as
	\begin{equation}
	\{M_{i \rightarrow k}, Q_{k \rightarrow i}\} = \mathcal{N}_{\text{flow}}(I_k, I_i).
	\end{equation}
	If $Q_{k\rightarrow i}(p) \leq \tau$, the propagated feature $F_{k \rightarrow i}(p)$ is inconsistent with the real feature $F_{i}(p)$. That is to say, $F_{k \rightarrow i}(p)$ is a bad approximation, which suggests updating with real feature $F_{i}(p)$.
	
	We consider a partial feature updating for non-key frames. Feature at frame $i$ is updated by
	\begin{equation}
	\hat{F}_i = U_{k \rightarrow i} \odot \mathcal{N}_{\text{feat}}(I_i) + (1-U_{k \rightarrow i}) \odot F_{k \rightarrow i},
	\label{eq.partial_update}
	\end{equation}
	where the updating mask $U_{k \rightarrow i}(p) = 1$ if $Q_{k \rightarrow i}(p) \leq \tau$, and $U_{k \rightarrow i}(p) = 0$, otherwise.
	In our implementation, we adopt a more economic way which recomputes feature $\hat{F}^{(n)}_i$ of layer $n$ from $\mathcal{N}^{(n)}_{\text{feat}}(\hat{F}^{(n-1)}_i)$, where $\hat{F}^{(n-1)}_i$ is the partially updated feature of layer $n-1$. Thus the partial feature updating can be calculated layer-by-layer. Considering varied resolution of feature maps in different layers, we use nearest neighbor interpolation for the updating mask.
	
	Following \cite{bengio2016bnn}, we use a straight-through estimator for the gradient $\frac{\partial U_{k \rightarrow i}(p)}{\partial Q_{k \rightarrow i}(p)} = -1$, if $|Q_{k \rightarrow i}(p) - \tau| \leq 1$, $\frac{\partial U_{k \rightarrow i}(p)}{\partial Q_{k \rightarrow i}(p)} = 0$, otherwise.
	Thus it is fully differentiable.
	We can regard $Q_{k \rightarrow i}(p)-\tau$ as a new valuable for the estimation of $Q_{k \rightarrow i}(p)$, since $\tau$ can be viewed as the bias of $Q_{k \rightarrow i}(p)$, which takes no effect to the estimate $Q_{k \rightarrow i}(p)$. For simplicity, we directly set $\tau = 0$ in this paper.
	
	To further improve the feature quality for non-key frames, feature aggregation is also utilized as similar as Eq.~\ref{eq.sparse_recursive_aggregation}:
	\begin{equation}
	\bar{F}_i = W_{k\rightarrow i} \odot \bar{F}_{k \rightarrow i} + W_{i\rightarrow i} \odot \hat{F}_i,
	\label{eq.partial_aggregation}
	\end{equation}
	where the weight is normalized by $W_{k\rightarrow i}(p)+W_{i\rightarrow i}(p)=1$ at every location $p$.
	
	\subsection{Temporally-adaptive Key Frame Scheduling}
	\label{sec.temporally_adaptive}
	
\begin{figure*}
		\begin{center}
			\includegraphics[width=1.0\linewidth]{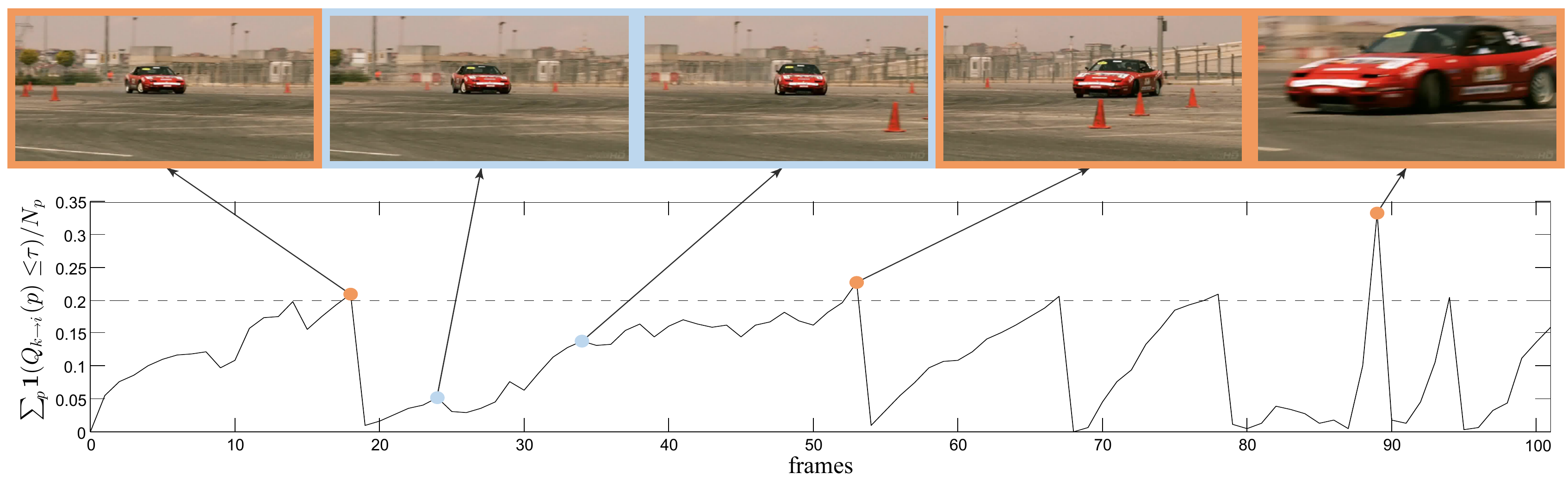}
		\end{center}
		\vspace{-0.5em}
		\caption{The area satisfying $Q_{k \rightarrow i}(p) \leq \tau$ on video frames, where the key frame scheduling in Eq.~\eqref{eq.iskey_funciton} is applied ($\gamma = 0.2$).}
		\vspace{-0.5em}
		\label{fig.adaptive_key_frame_example}\vspace{-0.5em}
	\end{figure*}

	Evaluating feature network $\mathcal{N}_{\text{feat}}$ only on sparse key frames is crucial for high speed.
	A naive key frame scheduling policy picks a key frame at a pre-fixed rate, \eg, every $l$ frames\cite{zhu2016dff}. A better key frame scheduling policy should be adaptive to the varying dynamics in the temporal domain. It can be designed based on the feature consistency indicator $Q_{k\rightarrow i}$:
	\begin{equation}
	key=is\_key(Q_{k \rightarrow i}).
	\end{equation}
	Here we designed a simple heuristic $is\_key$ function:
	{\small
		\begin{equation}
		is\_key(Q_{k \rightarrow i}) = [\frac{1}{N_p} \sum_p \mathbf{1}(Q_{k \rightarrow i}(p) \leq \tau) ] > \gamma
		\label{eq.iskey_funciton}
		\end{equation}}
	where $\mathbf{1}(\cdot)$ is the indicator function, $N_p$ is the number of all locations $p$.
	For any location $p$, $Q_{k \rightarrow i}(p) \leq \tau$ indicates changing appearance or large motion which will lead to bad feature propagation quality, if the area to recompute ($Q_{k \rightarrow i}(p) \leq \tau$) is larger than a portion $\gamma$ of all the pixels, the frame is marked as key.
	Figure.~\ref{fig.adaptive_key_frame_example} shows an example of the area satisfied $Q_{k \rightarrow i}(p) \leq \tau$ varying through time.
	Three orange points are examples of key frame selected by our $is\_key$ function, their appearance are clearly different.
	Two blue points are examples of non-key frame, their appearance indeed changed slightly compared with the preceding key frame. 
	
	To explore the potential and upper bound of key frame scheduling, we designed an oracle scheduling policy that exploits the ground-truth information. The experiment is performed with our proposed method, except for key frame scheduling policy. Given any frame $i$, both the detection results of picking frame $i$ as a key frame or non-key frame are computed, and the two mAP scores are also computed using ground truth.
	If picking it as a key frame results a higher mAP score, frame $i$ is marked as key.
	
	This oracle scheduling achieves a significantly better result, \ie, 80.9\% mAP score at 22.8 fps runtime speed. This indicates the importance of key frame scheduling and suggests that it is an important future working direction.
	
	\setlength{\tabcolsep}{8pt}
	\renewcommand{\arraystretch}{1}
	\begin{table*}[t]
		\begin{center}
			\begin{tabular}{l | l | l | l | l | l }
				\hline
				method & $is\_key(\cdot,\cdot)$ & key frame usage & $do\_aggr$ & $do\_spatial$ & accuracy$\leftrightarrow$speed\\
				\hline
				per-frame baseline (*) & all frames & N.A & $false$ & $false$ & none\\
				\hline
				Sparse Feature Propagation~\cite{zhu2016dff} & every $l$ frames & sparse, $1$ & $false$ & $false$ & $l$ \\
				Dense Feature Aggregation~\cite{zhu2017flow} & all frames & dense, $\geq 1$ & $true$ & $false$ & \#key frames \\
				\hline
				our method (c1) & every $l$ frames & sparse, recursive & $true$ & $false$ & $l$ \\
				our method (c2) & every $l$ frames & sparse, recursive & $true$ & $true$ & $l$, $\lambda$ \\
				\textbf{our method (c3)} & temporally-adaptive  & sparse, recursive & $true$ & $true$ & $\lambda, \gamma$ \\
				\hline
			\end{tabular}
		\end{center}
		\caption{All methods under a unified viewpoint.}
		\label{table.unified_methods}
	\end{table*}
	
	\subsection{A Unified Viewpoint}
	\label{sec.unified_viewpoint}
	All methods are summarized under a unified viewpoint.
	
	To efficiently compute feature maps, \emph{Spatially-adaptive Partial Feature Updating} (see Section~\ref{sec.spatially_adaptive}) is utilized.
	Although Eq.~\eqref{eq.partial_update} is only defined for non-key frames, it can be generalized to all frames.
	Given a frame $i$ and its preceding key frame $k$, Eq.~\eqref{eq.partial_update} is utilized, and summarized as
	\begin{equation}
	\hat{F}_i=PartialUpdate(I_i, F_k, M_{i \rightarrow k}, Q_{k \rightarrow i}).
	\label{eq.unified_partial_update}
	\end{equation}
	For key frames, $Q_{k\rightarrow i} = \mathbf{-\infty}$, propagated features $F_{k \rightarrow i}$ are always bad approximation of real features $F_i$, we should recompute feature $\hat{F}_i=\mathcal{N}_{\text{feat}}(I_i)$.
	For non-key frames, when $Q_{k\rightarrow i} = \mathbf{+\infty}$, propagated features $F_{k \rightarrow i}$ are always good approximation of true features $F_i$, we directly use the propagated feature from the preceding key frame $\hat{F}_i=F_{k \rightarrow i}$.
	
	To enhance the partially updated feature maps $\hat{F}_i$, feature aggregation is utilized.
	Although Eq.~\eqref{eq.sparse_recursive_aggregation} only defined \emph{Sparsely Recursive Feature Aggregation} for key frames, and Eq.~\eqref{eq.partial_aggregation} only defined feature aggregation for partially updated non-key frames. Eq.~\eqref{eq.sparse_recursive_aggregation} can be regarded as a degenerated version of Eq.~\eqref{eq.partial_aggregation}, supposing $i = k', \hat{F}_i = F_{k'}$.
	Thus feature aggregation is always performed as Eq.~\eqref{eq.partial_aggregation}, and summarized as
	\begin{equation}
	\bar{F}_i=\mathcal{G}(\bar{F}_k, \hat{F}_i, M_{i \rightarrow k}),
	\label{eq.unified_aggregation}
	\end{equation}
	
	To further improves the efficiency of feature computation, \emph{Temporally-adaptive Key Frame Scheduling} (see Section~\ref{sec.temporally_adaptive}) is also utilized.
	
	\paragraph{Inference}
	Algorithm~\ref{alg.inference_unified} summarizes the unified inference algorithm. Different settings result in different degenerated versions, and Table~\ref{table.unified_methods} presents all methods from the unified viewpoint. Our method (c3) integrates all the techniques and works best.
	
	If \emph{Temporally-adaptive Key Frame Scheduling} is adopted, and both options $do\_aggr$ and $do\_spatial$ are set as $true$, then it is the online version of our proposed method.
	Utilizing a naive key frame scheduling, \ie, pick a key every $l$ frame, and both options $do\_aggr$ and $do\_spatial$ set as $false$, the algorithm degenerates to Sparse Feature Propagation~\cite{zhu2016dff} when $l > 1$, and the per-frame baseline when $l = 1$.
	The algorithm would degenerate to Dense Feature Aggregation~\cite{zhu2017flow} under condition that $do\_aggr = true, key = true$, and $do\_spatial = false$ for all the frames (\ie, $l = 1$), and the unified feature aggregation on Line~\ref{alg:line:aggregate} is replaced by the dense aggregation in Eq.~\eqref{eq.feature_aggregation}.
	Among all options in Table~\ref{table.unified_methods}, a sparse key frame scheduling is crucial of fast inference, $do\_aggr=true$ and $do\_spatial=true$ is crucial for high accuracy.
	
	\paragraph{Training}
	
	All the modules in the entire architecture can be jointly trained. Due to memory limitation, in SGD, two nearby frames are randomly sampled in each mini-batch. The preceding frame is set as key, and the succeeding one is set as non-key, which are denoted as $I_k$ and $I_i$, respectively. 
	
	In the forward pass, feature network $\mathcal{N}_{\text{feat}}$ is applied on $I_k$ to obtain the feature maps $F_k$. Next, a flow network $\mathcal{N}_{\text{flow}}$ runs on the frames $I_i$, $I_k$ to estimate the 2D flow field $M_{i \rightarrow k}$ and the feature consistency indicator $Q_{k \rightarrow i}$. 
	Partially updated feature maps $\hat{F}_i$ is computed through Eq.~\eqref{eq.partial_update}, and then the aggregated current feature maps $\bar{F}_i$ is calculated through Eq.~\eqref{eq.partial_aggregation}.
	Finally, the detection sub-network $\mathcal{N}_{\text{det}}$ is applied on $\bar{F}_i$ to produce the result $y_i$.
	Loss function is defined as,
	\begin{equation}
	L=L_{det}(y_i) + \lambda \sum_{p} U_{k \rightarrow i}(p),
	\end{equation}
	where the updating mask $U_{k \rightarrow i}$ is defined in Eq.~\eqref{eq.partial_update}.
	The first term is the loss function for object detection, following the multi-task loss in Faster R-CNN~\cite{ren2015faster}, which consists of classification loss and bounding box regression loss together. The second term enforces a constraint on the size of areas to be recomputed, and $\lambda$ controls the speed-accuracy trade off.
	
	During training, we enforce that $U_{k \rightarrow i} = \mathbf{0}$ and $U_{k \rightarrow i} = \mathbf{1}$ with $1/3$ probability, respectively, to encourage good performance for both cases of propagating feature and recomputing feature from scratch.
	For methods without using partial feature updating, training does not change and $Q_{k \rightarrow i}$ is simply ignored during inference. Thus, a unified single training strategy is used.
	
	\subsection{Network Architecture}
	
	We introduce the incarnation of different sub-networks in our proposed model.
	
	\textbf{Flow network.} We use FlowNet~\cite{dosovitskiy2015flownet} (``simple" version). It is pre-trained on the Flying Chairs dataset~\cite{dosovitskiy2015flownet}. It is applied on images of half resolution and has an output stride of 4. As the feature network has an output stride of 16 (see below), the flow field is downscaled by half to match the resolution of the feature maps. An additional randomly initialized 3x3 convolution is added to predict the feature propagability indicator, which shares feature with the last convolution of the FlowNet. 
	
	\textbf{Feature network.} We adopt the state-of-the-art ResNet-101~\cite{he2016deep} as the feature network. The ResNet-101 model is pre-trained on ImageNet classification.
	We slightly modify the nature of ResNet-101 for object detection. We remove the ending average pooling and the fc layer, and retain the convolution layers. To increase the feature resolution, following the practice in~\cite{chen2014semantic,dai2016rfcn}, the effective stride of the last block is changed from 32 to 16. Specially, at the beginning of the last block (``conv5" for both ResNet-101), the stride is changed from 2 to 1. To retain the receptive field size, the dilation of the convolutional layers (with kernel size $>$ 1) in the last block is set as 2. Finally, a randomly initialized $3 \times 3$ convolution is applied on top to reduce the feature dimension to 1024.
	
	\textbf{Detection network.} We use state-of-the-art R-FCN~\cite{dai2016rfcn} and follow the design in~\cite{zhu2016dff}. On top of the 1024-d feature maps, the RPN sub-network and the R-FCN sub-network are applied, which connect to the first 512-d and the last 512-d features respectively.  9 anchors (3 scales and 3 aspect ratios) are utilized in RPN, and 300 proposals are produced on each image. The position-sensitive score maps in R-FCN are of $7 \times 7$ groups.
	
	\begin{algorithm}[t]
		\caption{The unified flow-based inference algorithm for video object detection.}
		\small
		\begin{algorithmic}[1] 
			\State \textbf{input}: video frames $\{I_i\}$
			
			\State $k=0$                            \Comment{initialize key frame}
			\State $F_0 = \mathcal{N}_{\rm feat}(I_0)$
			\State $y_0 = \mathcal{N}_{\rm det}(F_0)$
			\If{$do\_aggr$}
			\State $\bar{F}_0 = F_0$
			\EndIf
			
			\For{$i=1$ \textbf{to} $\infty$}
			\State $\{M_{i \rightarrow k}, Q_{k \rightarrow i}\} = \mathcal{N}_{\text{flow}}(I_k, I_i)$ \Comment{evaluate flow network}
			\State $key=is\_key(Q_{k \rightarrow i})$ \Comment{key frame scheduling}
			\If{$key$}
			\State $Q_{k \rightarrow i}=\mathbf{-\infty}$ \Comment{need computing feature from scratch}
			\ElsIf{$do\_spatial$}
			\State $Q_{k \rightarrow i}$ unchanged \Comment{need partially updating}
			\Else
			\State $Q_{k \rightarrow i}=\mathbf{+\infty}$ \Comment{suppose always good quality, propagate}
			\EndIf
			
			\State $\hat{F}_i=PartialUpdate(I_i, F_k, M_{i \rightarrow k}, Q_{k \rightarrow i})$ \Comment{partially update}
			
			\If{$do\_aggr$}
			\State $\bar{F}_i=\mathcal{G}(\bar{F}_k, \hat{F}_i, M_{i \rightarrow k})$ \Comment{recursively aggregate} \label{alg:line:aggregate}
			\State $y_i=\mathcal{N}_{\rm det}(\bar{F}_i)$
			\Else
			\State $y_i=\mathcal{N}_{\rm det}(\hat{F}_i)$
			\EndIf
			
			\If{$key$} \Comment{update the most recent key frame}
			\State $k=i$
			\EndIf
			
			\EndFor
			\State \textbf{output}: detection results $\{y_{i}\}$
		\end{algorithmic}
		\label{alg.inference_unified}
	\end{algorithm}
	
	\section{Related Work}
	
	\textbf{Speed/accuracy trade-offs in object detection.}
	As summarized in \cite{huang2016speed}, speed/accuracy trade-off of modern detection systems can be achieved by using different feature networks~\cite{simonyan2015very,szegedy2015going,he2016deep,szegedy2016inception,xie2017resnext,huang2016densely,chollet2016xception,howard2017mobilenets,zhang2017shufflenet} and detection networks~\cite{girshick2014rich,he2014spatial,girshick2015fast,ren2015faster,dai2016rfcn,lin2016fpn,he2017mask,dai2017deformable,liu2016ssd,redmon2016you,redmon2016yolo9000,lin2017focal}, or varying some critical parameters such as image resolution, box proposal number.
	PVANET~\cite{kim2016pvanet} and YOLO~\cite{redmon2016you} even design specific feature networks for fast object detection. By applying several techniques (\eg batch normalization, high resolution classifier, fine-grained features and multi-scale training), YOLO9000~\cite{redmon2016yolo9000} achieves higher accuracy meanwhile keep the high speed.
	
	Since our proposed method only considers how to compute higher quality feature faster by using temporal information, and is not designed for any specific feature networks and detection networks, such techniques are also suitable for our proposed method.
	
	\textbf{Video object detection.}
	Existing object detection methods incorporating temporal information in video can be separated into box-level methods ~\cite{kang2016object, kang2016tcnn, han2016seqnms, lee2016multi, kang2017tpn, Feichtenhofer17DetectTrack} and feature-level methods ~\cite{zhu2016dff, zhu2017flow} (both are flow-based methods and introduced in Section~\ref{sec.method_two_baselines}).
	
	Box-level methods usually focus on how to improve detection accuracy considering temporary consistency within a tracklet.
	T-CNN~\cite{kang2016tcnn, kang2016object} first propagates predicted bounding boxes to neighboring frames according to pre-computed optical flows, and then generates tubelets by applying tracking algorithms. Boxes along each tubelet will be re-scored based on the tubelet classification result.
	Seq-NMS~\cite{han2016seqnms} constructs sequences along nearby high-confidence bounding boxes from consecutive frames. Boxes of the sequence are re-scored to the average confidence, other boxes close to this sequence are suppressed. 
	MCMOT~\cite{lee2016multi} formulates the post-processing as a multi-object tracking problem, and finally tracking confidence are used to re-score detection confidence.
	TPN~\cite{kang2017tpn} first generates tubelet proposals across multiple frames ($\leq$ 20 frames) instead of bounding box proposals in a single frame, and then each tubelet proposal is classified into different classes by a LSTM based classifier.
	D\&T~\cite{Feichtenhofer17DetectTrack} simultaneously outputs detection boxes and regression based tracking boxes with a single convolutional neural networks, and detection boxes are linked and re-scored based on tracking boxes.
	
	Feature-level methods usually use optical flow to get pixel-to-pixel correspondence among nearby frames. Although feature-level methods are more principle and can further incorporate with box-level methods, they suffer from inaccurate optical flow.
	Still ImageNet VID 2017 winner is powered by feature-level methods DFF~\cite{zhu2016dff} and FGFA~\cite{zhu2017flow}.
	Our proposed method is also a feature-level method, which introduces \emph{Spatially-adaptive Partial Feature Updating} to fix the inaccurate feature propagation caused by inaccurate optical flow.
	
	\section{Experiments}\label{sec.experiment}
	
	ImageNet VID dataset~\cite{russakovsky2015imagenet} is a prevalent large-scale benchmark for video object detection. Following the protocols in~\cite{kang2016tcnn,lee2016multi}, model training and evaluation are performed on the 3,862 video snippets from the training set and the 555 snippets from the validation set, respectively. The snippets are fully annotated, and are at frame rates of 25 or 30 fps in general. There are 30 object categories, which are a subset of the categories in the ImageNet DET dataset.
	
	During training, following \cite{kang2016tcnn,lee2016multi}, both the ImageNet VID training set and the ImageNet DET training set (only the same 30 categories as in ImageNet VID) are utilized.
	SGD training is performed.
	Each mini-batch samples one image from either ImageNet VID or ImageNet DET datasets, at 1 : 1 ratio.
	120K iterations are performed on 4 GPUs, with each GPU holding one mini-batch. The learning rates are $10^{-3}$ and $10^{-4}$ in the first 80K and in the last 40K iterations, respectively. 
	In both training and inference, the images are resized to a shorter side of 600 pixels for the image recognition network, and a shorter side of 300 pixels for the flow network. Experiments are performed on a workstation with Intel E5-2670 v2 CPU 2.5GHz and Nvidia K40 GPU.

	\subsection{Evaluation under a Unified Viewpoint}
	
\begin{figure}
		\begin{center}
			\includegraphics[width=0.9\linewidth]{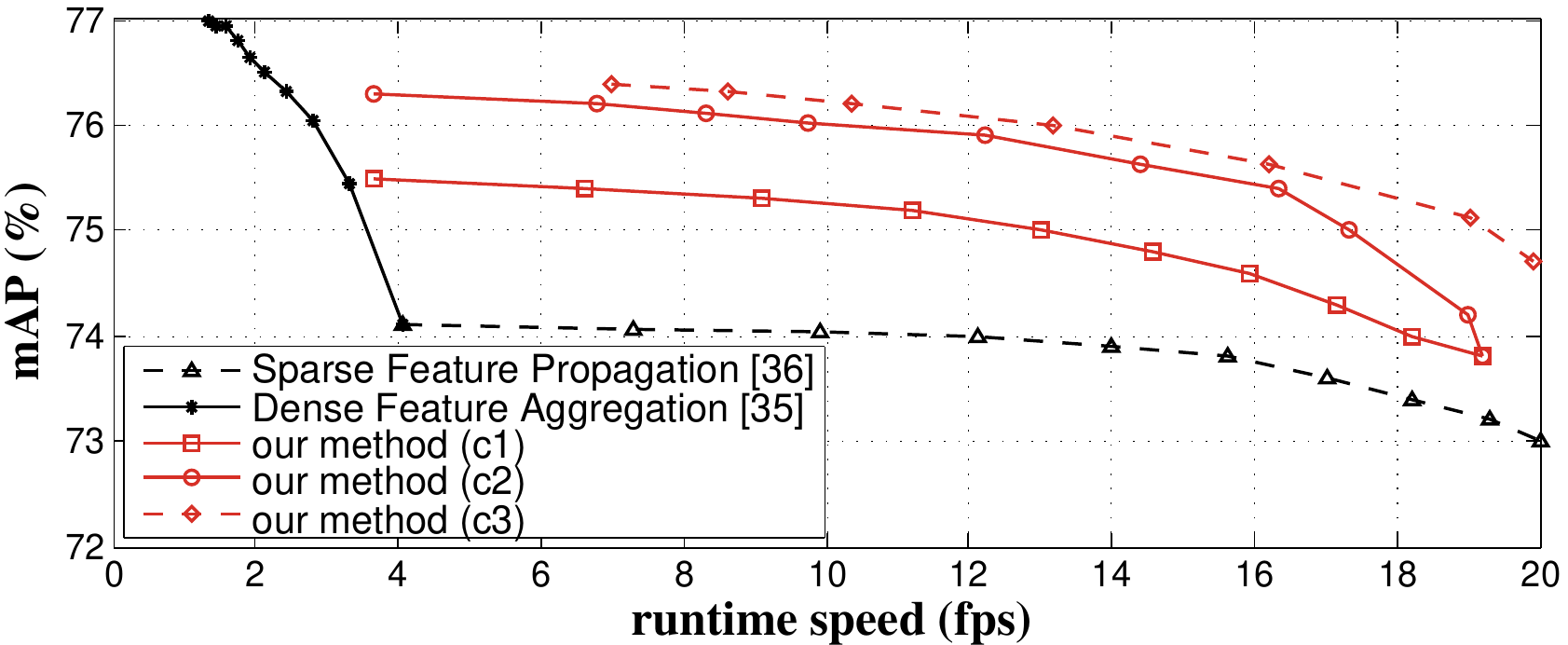}
		\end{center}
		\caption{Speed-accuracy trade-off curves for methods in Table~\ref{table.unified_methods}.}
		\label{fig.detection_fps_map_overall}\vspace{-0.5em}
	\end{figure}

	Overall comparison results are shown in Figure~\ref{fig.detection_fps_map_overall}.
	
	Sparse Feature Propagation~\cite{zhu2016dff} is a degenerated version in Algorithm~\ref{alg.inference_unified} (see Table~\ref{table.unified_methods}). By varying key frame duration $l$ from 1 to 10, it can achieve $5\times$ speedup with moderate accuracy loss (within $1\%$).
	
	Similarly, for Dense Feature Aggregation~\cite{zhu2017flow}, by varying the temporal window to be aggregated from $\pm 1$ to $\pm 10$ frames, it improves mAP score by $2.9\%$ but is $3 \times$ slower than per-frame baseline.
	
	For our method (c1), key frames are picked once every $l$ frames ($l = 1 \sim 10$ frames).
	Compared with Sparse Feature Propagation~\cite{zhu2016dff}, the only difference is $do\_aggr$ set as $true$ instead of $false$, which leads to almost $1\%$ improvement in mAP score with the same speedup.
	It recursively aggregates feature maps on sparse key frames, and the aggregated feature maps are propagated to non-key frames (see Eq.~\eqref{eq.sparse_recursive_aggregation}).
	Compared with Dense Feature Aggregation~\cite{zhu2017flow}, recursive aggregation is performed only on sparse key frames instead of dense feature aggregation performed on every frames, which leads to $10 \times$ speedup with $2\%$ accuracy loss.
	Compared with per-frame baseline, it achieves $1\%$ higher accuracy and $3 \times$ faster speed.
	
	Our method (c2) extends our method (c1) by setting $do\_spatial$ as $true$ instead of $false$.
	It can further utilize rich appearance information from nearby frames with negligible computation burden.
	Compared with Sparse Feature Propagation~\cite{zhu2016dff}, it improves mAP score with almost $2\%$ and keeps the same high speed.
	Compared with Dense Feature Aggregation~\cite{zhu2017flow}, it can speed up $9 \times$ with $1\%$ accuracy loss.
	Compared with per-frame baseline, this version results $1.8\%$ higher accuracy with $3 \times$ speedup and $1.4\%$ higher accuracy with $4\times$ speedup.
	
	Our method (c3) further extends our method (c2) by utilizing a temporally-adaptive key frame scheduling instead of a pre-fixed key frame duration.
	$\gamma$ in Eq.~\eqref{eq.iskey_funciton} is fixed as 0.2.
	Compared with our method (c2), it further improves detection accuracy with $0.5\% \sim 1\%$ when high runtime speed is demanded.
	Compared with Sparse Feature Propagation~\cite{zhu2016dff}, it improves mAP score with nearly $2\%$ at all runtime speed.
	Compared with per-frame baseline, this version results $1\%$ higher accuracy with $4.75 \times$ speedup.
	
	\subsection{Ablation Study}
	We conduct ablation study for three different options of our method. The detailed setting is shown in Table~\ref{table.unified_methods}. All of three options use sparsely recursive feature aggregation for key frame, and then propagate the aggregated features to non-key frames, \ie, $do\_aggr = true$. The difference among them is key-frame scheduling and whether partial feature updating is used or not.
	
\begin{figure}
		\begin{center}
			\includegraphics[width=0.9\linewidth]{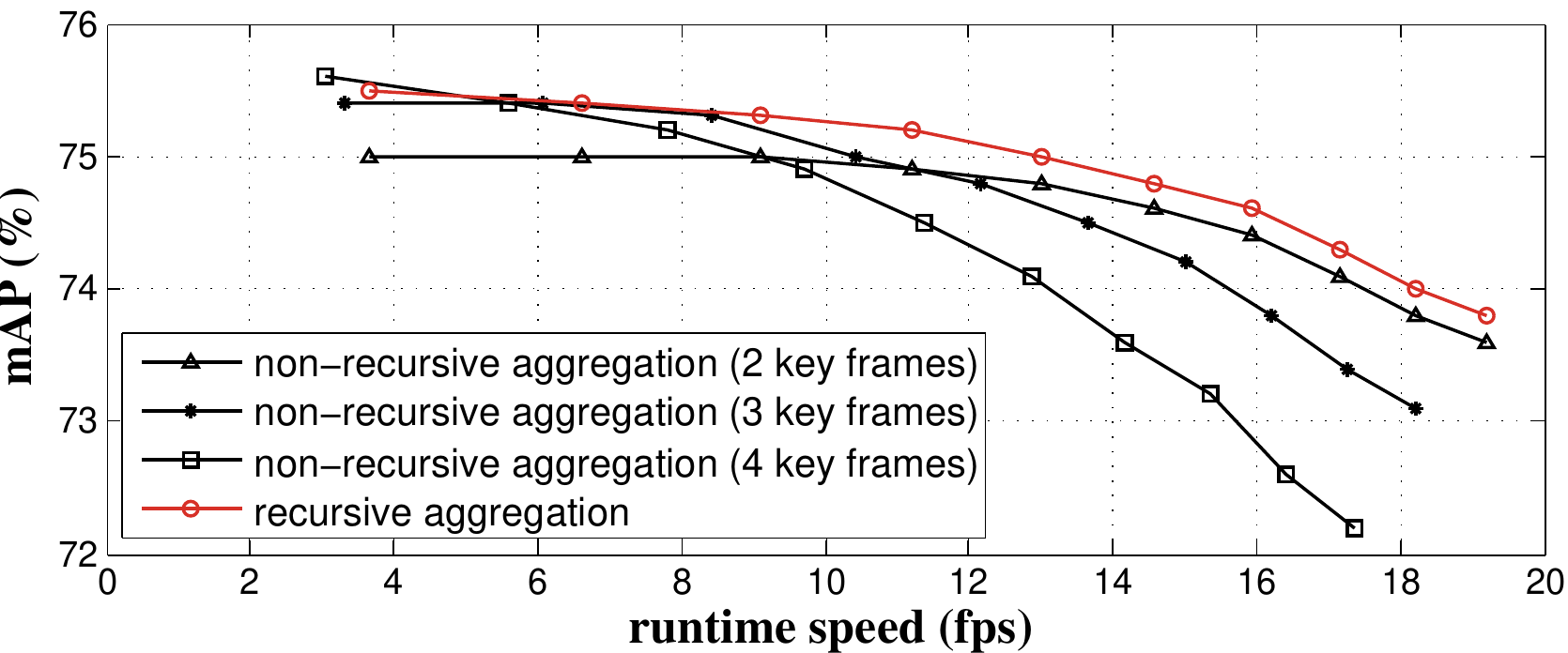}
		\end{center}
		\caption{Speed-accuracy trade-off curves for our method (c1) and its non-recursive aggregation variants.}
		\label{fig.detection_fps_map_c1}\vspace{-0.5em}
	\end{figure}

	\textbf{Our method (c1)} We evaluate the effect of recursive feature aggregation compared with non-recursive aggregation (\ie, dense aggregation) on sparse key frames. Here, we use several variant numbers of key frames for non-recursive aggregation. Results are shown in Figure~\ref{fig.detection_fps_map_c1}. For non-recursive aggregation methods, aggregating more key frames is better when runtime speed is slow. Moreover, when aggregating more than 2 key frames, accuracy descends quickly. It is caused by feature inconsistency from propagated key frames with large key frame duration $l$, which is on the demand for high runtime speed. 
	Recursive aggregation can solve this problem well by only considering two key frames in aggregation. More important, the aggregated feature theoretically contains all historical information of previous key frames. So the aggregation no longer needs more key frames (larger than 2 frames). As we can see, recursive aggregation surpasses the non-recursive aggregation at almost all runtime speed.
	
\begin{figure}
		\begin{center}
			\includegraphics[width=0.9\linewidth]{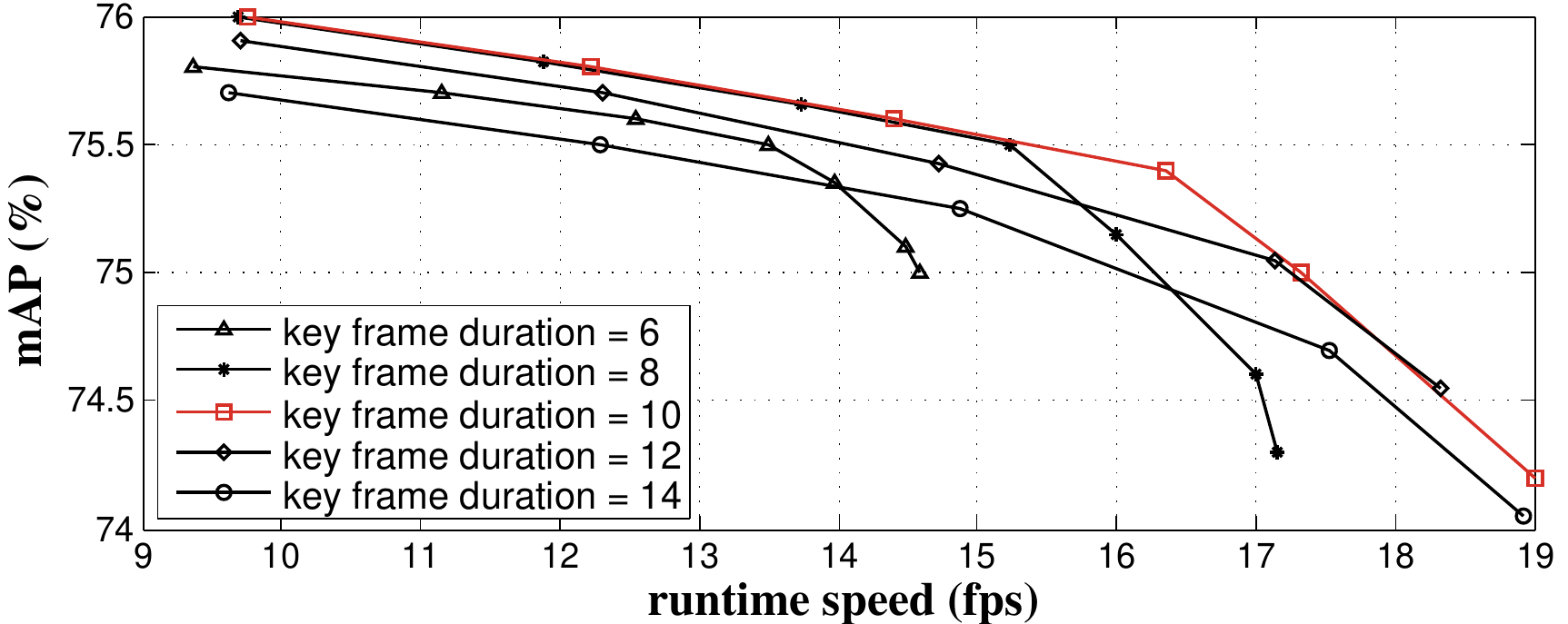}
		\end{center}
		\caption{Speed-accuracy trade-off curves for our method (c2), and each curve shares a fixed key frame duration $l$.}
		\label{fig.detection_fps_map_c2}\vspace{-0.5em}
	\end{figure}

\begin{figure}
		\begin{center}
			\includegraphics[width=0.9\linewidth]{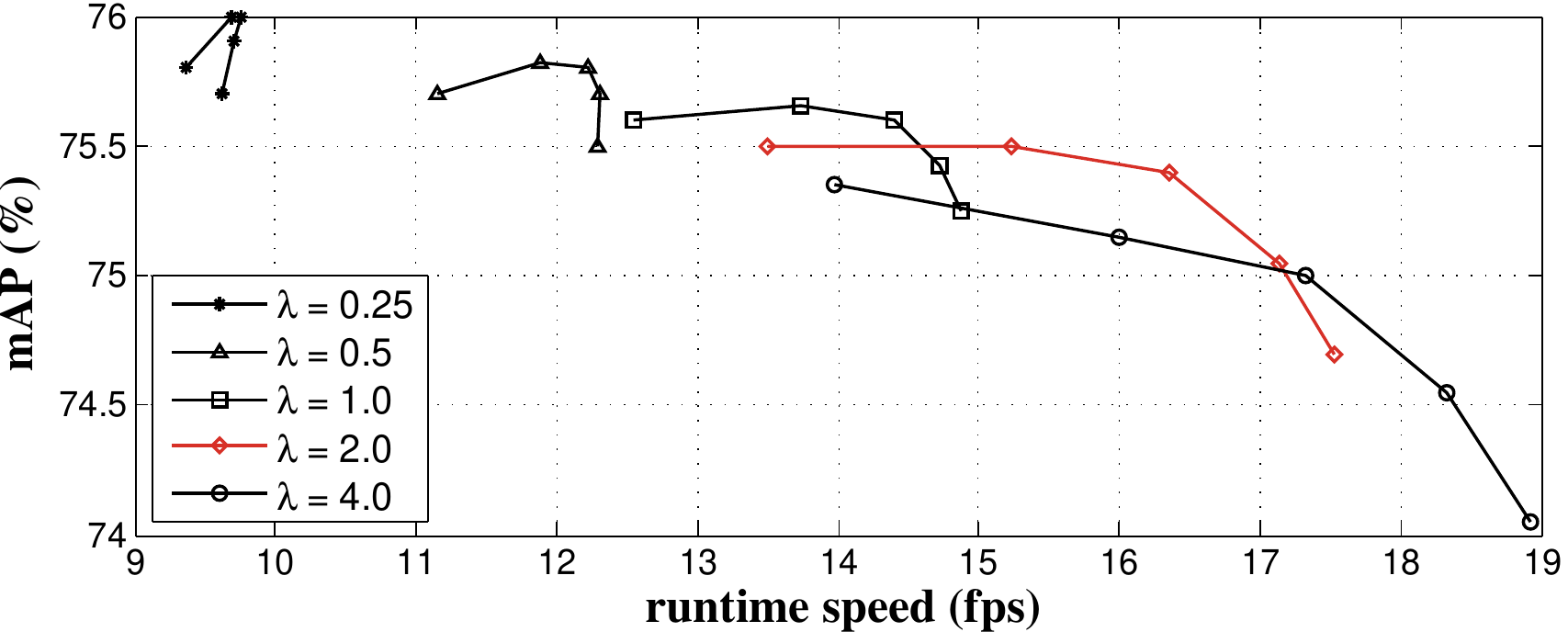}
		\end{center}
		\caption{Speed-accuracy trade-off curves for our method (c2), and each curve shares a fixed partially updating coefficient $\lambda$.}
		\label{fig.detection_fps_map_c2_2}\vspace{-0.5em}
	\end{figure}

	\textbf{Our method (c2)} We evaluate the effect of partially updating coefficient $\lambda$ and key frame duration $l$, which actually controls the speed-accuracy trade-off.
	Figure~\ref{fig.detection_fps_map_c2} shows the results with varying $\lambda$ and fixed $l$.
	Key frame duration $l = 10$ achieves the best speed-accuracy trade-off.
	Small $l$ leads to redundancy between two consecutive key frames, which is not useful for recursive aggregation, thus results in a little accuracy loss.
	Large $l$ leads to highly diverse feature response between two consecutive key frames, which is also not helpful.
	Figure~\ref{fig.detection_fps_map_c2_2} shows the results with varying $l$ and fixed $\lambda$.
	Partially updating coefficient $\lambda = 2.0$ achieves the best speed-accuracy trade-off.
	Small $\lambda$ implies very large recomputed area, and always gives low runtime speed regardless of key frame duration.
	High $\lambda$ implies very small recomputed area, which does not fully exploit the strength of partially updating.

\begin{figure}
		\begin{center}
			\includegraphics[width=0.9\linewidth]{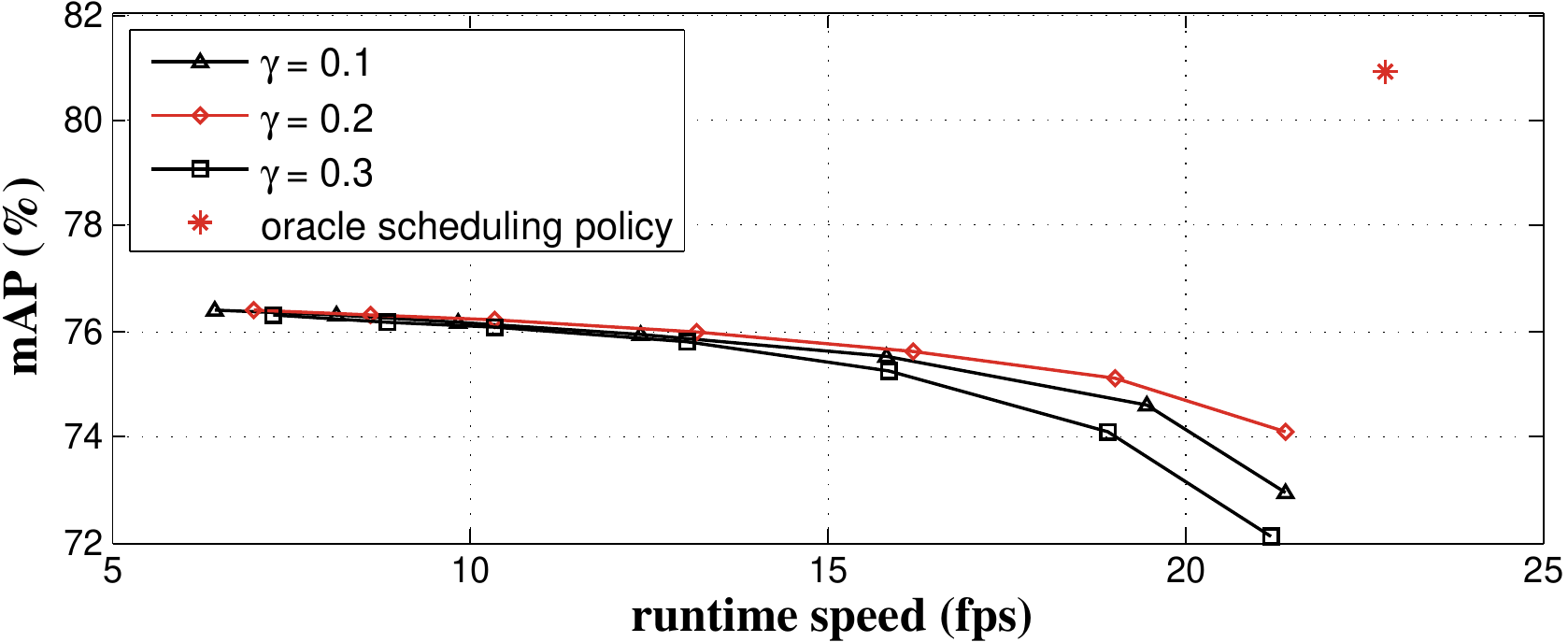}
		\end{center}
		\caption{Speed-accuracy trade-off curves for our method (c3) with different $\gamma$.}
		\label{fig.detection_fps_map_c3}\vspace{-0.5em}
	\end{figure}

	\textbf{Our method (c3)} We compare our \emph{Temporally-adaptive Key Frame Scheduling} with different $\gamma$ (see Eq.~\eqref{eq.iskey_funciton}), the results are showed in Figure~\ref{fig.detection_fps_map_c3}.
	Different $\gamma$s result almost the same performance when runtime speed is slow. $\gamma=0.2$ results best speed-accuracy trade off when high runtime speed is demanded.
	The oracle key frame scheduling policy (described in Section.~\ref{sec.temporally_adaptive}) achieves an incredibly better results.
	
\begin{figure}
		\begin{center}
			\includegraphics[width=0.9\linewidth]{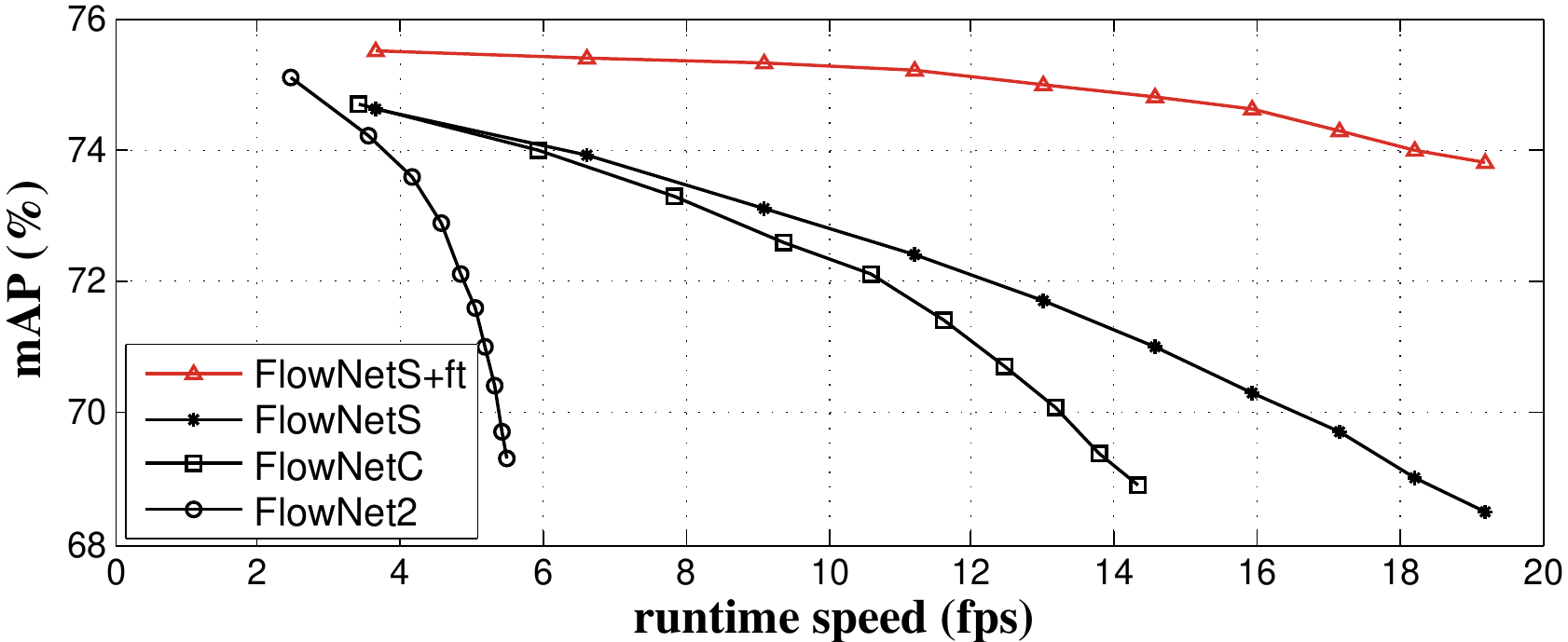}
		\end{center}
		\caption{Speed-accuracy trade-off curves for our method (c1) with different flow networks. `FlowNetS+ft' stands for FlowNetS jointly trained within our proposed method. Other flow networks are used without joint training.}
		\label{fig.detection_fps_map_c1_flownets}\vspace{-0.5em}
	\end{figure}

	\textbf{Different flow networks} We also evaluated different flow networks (including FlowNetS, FlowNetC and FlowNet2~\cite{ilg2016flownet2}) for our proposed method. Results are showed in Figure.~\ref{fig.detection_fps_map_c1_flownets}.
	FlowNetS results best speed-accuracy trade-off, this is because fast inference of flow network is the key to speedup in our proposed method.
	With joint training, FlowNetS can achieve significantly better results, which is consistent with \cite{zhu2016dff, zhu2017flow}.
	
\begin{figure}
		\begin{center}
			\includegraphics[width=0.9\linewidth]{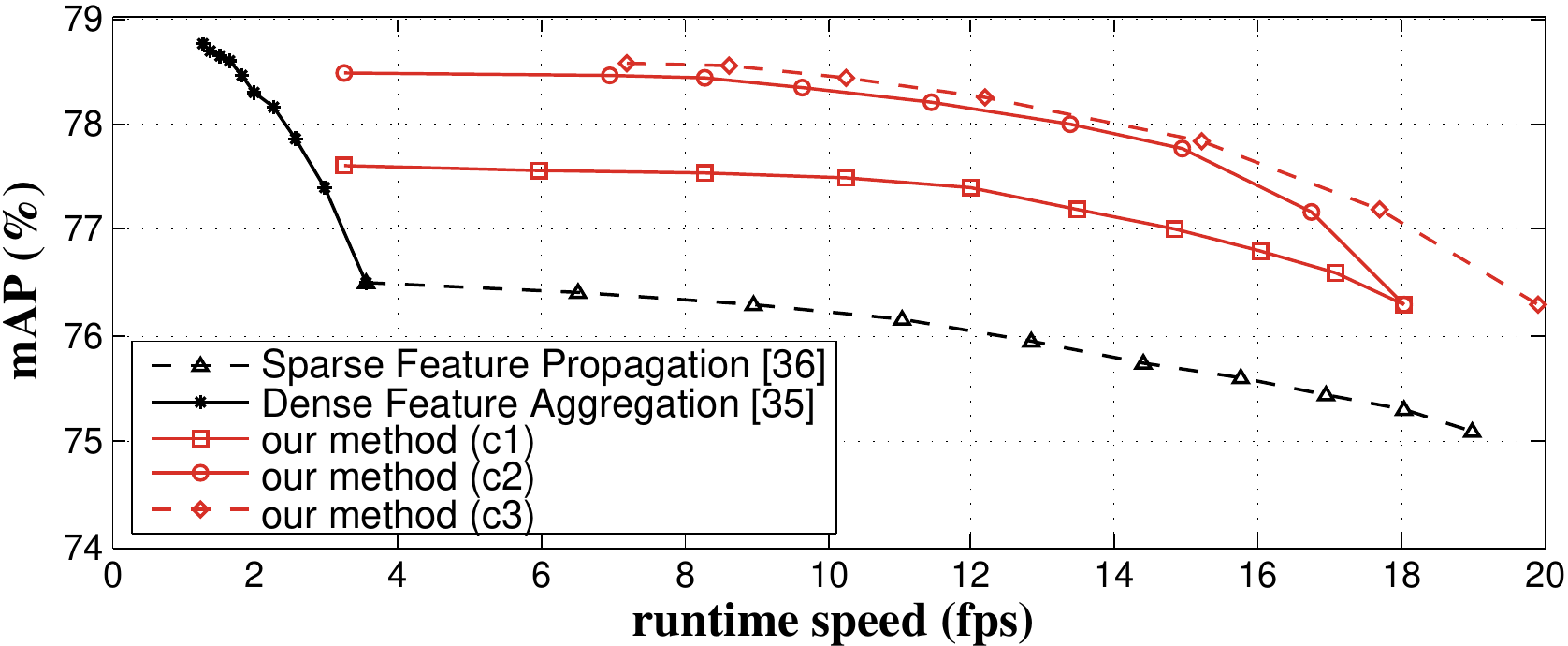}
		\end{center}
		\caption{Speed-accuracy trade-off curves for all methods in Table~\ref{table.unified_methods} combined with Deformable R-FCN.}
		\label{fig.detection_fps_map_overall_dcn}\vspace{-0.5em}
	\end{figure}

	\textbf{Deformable R-FCN~\cite{dai2017deformable}} We further replace the detection system with Deformable R-FCN, which is slightly slower than the original R-FCN but much more accurate. Results are showed in Figure.~\ref{fig.detection_fps_map_overall_dcn}. Our proposed method works well, and achieves $77.8\%$ mAP score at 15.2 fps runtime speed, better than ImageNet VID 2017 winner ($76.8\%$ mAP score at 15.4 fps runtime speed~\cite{deng2017vid}).
	
	\subsection{Comparison with State-of-the-art Methods}
	
	We further compared with several state-of-the-art methods \& systems for object detection from video, with reported results on ImageNet VID validation. It is worth mentioning that different recognition networks, object detectors, and post processing techniques are utilized in different approaches. Thus it is hard to draw a fair comparison.
	
	Table~\ref{tab.comparison_sota} presents the results. For our method, we reported results by picking two operational points on curve ``our method (c3)" from Figure~\ref{fig.detection_fps_map_overall_dcn}. The mAP score is 78.6\% at a runtime of 13.0 / 8.6 fps on Titan X / K40. The mAP score slightly decrease to 77.8\% at a faster runtime of 22.9 / 15.2 fps on Titan X / K40. As a comparison, TPN~\cite{kang2017tpn} gets an mAP score of 68.4\% at a runtime of 2.1 fps on Titan X. In the latest paper of D\&T~\cite{Feichtenhofer17DetectTrack}, an mAP score of 75.8\% is obtained at a runtime of 7.8 fps on Titan X. Sequence NMS~\cite{han2016seqnms} can be applied to D\&T to further improve the performance, which can also be applied in our approach. We also compared with the winning entry~\cite{deng2017vid} of ImageNet VID challenge 2017, which is also based on sparse feature propagation~\cite{zhu2016dff} and dense feature aggregation~\cite{zhu2017flow}. It gets an mAP score of 76.8\% at a runtime of 15.4 fps on Titan X. It is heavily-engineered and the implementation details are unreported. Our method is more principled, and achieves better performance in terms of both accuracy and speed.

	\setlength{\tabcolsep}{2pt}
	\renewcommand{\arraystretch}{1.2}
	\begin{table}
		\small
		\begin{center}
			\begin{tabular}{l|c|c|c}
				\hline
				method &  feature network	& mAP (\%)  & \tabincell{c}{runtime (fps) \\ (TitanX/K40)} \\
				\hline\hline
				\multirow{2}{*}{Ours} & \footnotesize \multirow{2}{*}{ResNet-101+DCN} & \textbf{78.6} & 13.0 / 8.6 \\
				& & 77.8 & 22.9 / 15.2 \\
				\hline
				\footnotesize TPN~\cite{kang2017tpn} & \footnotesize GoogLeNet & 68.4 & 2.1 / - \\
				\hline 
				\footnotesize D\&T~\cite{Feichtenhofer17DetectTrack} & \footnotesize ResNet-101 & 75.8 & 7.8 / - \\
				\hline 
				\footnotesize \tabincell{c}{ImageNet VID \\ 2017 winner~\cite{deng2017vid}} & \footnotesize {ResNet-101} & {76.8} & {- / 15.4} \\
				\hline 
			\end{tabular}
		\end{center}
		\caption{Comparison with state-of-the-art methods.}
		\label{tab.comparison_sota}
	\end{table}

	{
		\small
		\bibliographystyle{ieee}
		\bibliography{egbib}
	}
	
\end{document}